\documentclass[letterpaper, 10 pt, conference]{ieeeconf}  

\IEEEoverridecommandlockouts                              

\usepackage[OT1]{fontenc}

\usepackage{graphics} 
\usepackage{subfigure}

\usepackage{algorithm}
\usepackage[noend]{algpseudocode}
\newcommand*\Let[2]{\State #1 $\gets$ #2}

\usepackage{amsmath} 
\usepackage{amssymb}  
\usepackage{mathtools}

\usepackage{amsthm}

\newtheoremstyle{assumptionstyle}
  {0pt}  
  {0pt}  
  {\itshape}  
  {1em}  
  {\bfseries}  
  {.}  
  {1em}  
  {}  
\theoremstyle{assumptionstyle}
\newtheorem{assumption}{Assumption}

\newtheoremstyle{definitionstyle}
  {0pt}  
  {0pt}  
  {\normalfont}  
  {1em}  
  {\bfseries}  
  {.}  
  {1em}  
  {}  

\theoremstyle{definitionstyle}
\newtheorem{definition}{Definition}

\renewcommand{\vec}[1]{\mathbf{\MakeLowercase{#1}}}
\newcommand{\Mat}[1]{\mathbf{\MakeUppercase{#1}}}

\let\originalleft\left
\let\originalright\right
\renewcommand{\left}{\mathopen{}\mathclose\bgroup\originalleft}
\renewcommand{\right}{\aftergroup\egroup\originalright}

\DeclareMathOperator{\E}{E}
\newcommand{\EB}[1]{\ensuremath{\E \left[ #1 \right]}}

\DeclareMathOperator*{\argmin}{argmin}
\renewcommand{\P}{\operatorname{\mathbf{P}}}
\newcommand{\eye}{\Mat{I}}
\newcommand{\T}{\intercal}

\DeclarePairedDelimiter\norm{\lVert}{\rVert}
\DeclarePairedDelimiter\abs{\lvert}{\rvert}

\usepackage{ifthen}
\newcommand*{\indA}[3][]{%
    \ifthenelse{\equal{#1}{}}{#3_{#2}}{#3_{#2 | #1}}%
}
\newcommand*{\indki}[3][k]{%
    \ifthenelse{\equal{#1}{}}{\indA{#2}{#3}}{%
        \ifthenelse{\equal{#2}{}}{\indA[#1]{#1}{#3}}{\indA[#1]{#1 + #2}{#3}}
        }%
}
\newcommand*{\indk}[2]{\indki[]{#1}{#2}}
\newcommand*{\indTraj}[4][]{\indA[#1]{[#2,#3]}{#4}}

\newcommand*{\xA}[2][]{%
    \indA[#1]{#2}{x}%
}
\newcommand*{\xki}[2][k]{%
    \ifthenelse{\equal{#1}{}}{\xA{#2}}{%
        \ifthenelse{\equal{#2}{}}{\xA[#1]{#1}}{\xA[#1]{#1 + #2}}
        }%
}
\newcommand*{\xk}[1]{\xki[]{#1}}
\newcommand*{\xTraj}[3][]{\indTraj[#1]{#2}{#3}{x}}

\newcommand*{\uA}[2][]{%
    \indA[#1]{#2}{u}%
}
\newcommand*{\uki}[2][k]{%
    \ifthenelse{\equal{#1}{}}{\uA{#2}}{%
        \ifthenelse{\equal{#2}{}}{\uA[#1]{#1}}{\uA[#1]{#1 + #2}}
    }%
}
\newcommand*{\uk}[1]{\uki[]{#1}}
\newcommand*{\uTraj}[3][]{\indTraj[#1]{#2}{#3}{u}}

\newcommand*{\obsA}[2][]{%
    \indA[#1]{#2}{\gls{obs}}%
}
\newcommand*{\obski}[2][k]{%
    \ifthenelse{\equal{#1}{}}{\obsA{#2}}{\obsA[#1]{#1 + #2}}%
}
\newcommand*{\obsk}[1]{\obski[]{#1}}
\newcommand*{\obsTraj}[3][]{\indTraj[#1]{#2}{#3}{\gls{obs}}}

\newcommand*{\nk}[1]{\indk{#1}{z}}

\newcommand*{\nTraj}[3][]{\indTraj[#1]{#2}{#3}{z}}

\renewcommand*{\xk}[1]{\xki[k]{#1}}
\renewcommand*{\uk}[1]{\uki[k]{#1}}
\renewcommand*{\obsk}[1]{\obski[k]{#1}}
\renewcommand*{\indk}[2]{\indki[k]{#1}{#2}}

\usepackage[acronyms, style=alttree, nonumberlist, nogroupskip, nopostdot, shortcuts]{glossaries} 

\newacronym[description=Optimal control problem]{ocp}{OCP}{Optimal Control Problem}
\newacronym[description=Model predictive control]{mpc}{MPC}{Model Predictive Control}
\newacronym[description=Stochastic model-predictive control]{smpc}{SMPC}{Stochastic Model Predictive Control}
\newacronym[description=Robust model-predictive control]{rmpc}{RMPC}{Robust Model Predictive Control}
\newacronym[description=Advanced Driver Assistance System]{adas}{ADAS}{Advanced Driver Assistance System}
\newacronym[description=Ego vehicle]{ev}{EV}{Ego Vehicle}
\newacronym[description=Traffic participant]{tp}{TP}{Traffic Participant}
\newacronym[description=Probability density function]{pdf}{PDF}{Probability Density Function}
\newacronym[description=Cumulative distribution function]{cdf}{CDF}{Cumulative Distribution Function}
\newacronym[description=Closed-loop]{cl}{CL}{Closed-Loop}
\newacronym[description=Open-loop]{ol}{OL}{Open-Loop}
\newacronym[description=Certainty equivalent controller]{cec}{CEC}{Certainty Equivalent Controller}
\newacronym[description=Certainty equivalence principle]{cep}{CEP}{Certainty Equivalence Principle}
\newacronym[description=Quadratic Programming]{qp}{QP}{Quadratic Programming}
\newacronym{method_uts}{UT-\hspace{0pt}Funnel}{Uncertainty Target Funnel}

\newcommand{\Symbols}{List of Symbols}
\newglossary{symbols}{sym}{sbl}{\Symbols}
\newglossaryentry{ts}{type=symbols,
	sort={other},
	name={\ensuremath{T_s}},
	description={sampling time}
}
\newglossaryentry{ref}{type=symbols,
	sort={aad},
	name={\ensuremath{R \in \mathbb{R}^n}},
    text={\ensuremath{R}},
	description={reference with size $n$}
}
\newglossaryentry{obs}{type=symbols,
	sort={aae},
	name={\ensuremath{b}},
    text={\ensuremath{b}},
	description={environment belief (probability density function)}
}
\newglossaryentry{parameter-p}{type=symbols,
	sort={parameter},
	name={\ensuremath{\rho}},
	description={parameter for the Uncertainty Dependent Target Funnel method determining the target funnel size}
}

\newglossaryentry{quantile}{type=symbols, 
	sort={ad},
	name={\ensuremath{x_{(p)}}},
	description={element-wise $p$-th quantile of random variable $x$:\\$x_{(p)} = \min \left\{ 
		\begin{pmatrix}
			z_1 \\
			\vdots
		\end{pmatrix}	
	\mathop{:}
	\begin{pmatrix}
		\P(x_1 \leq z_1) \\
		\vdots
	\end{pmatrix} \geq p \right\}$}
}

\newglossary{scenarios}{sce}{sce}{\Scenarios}
\def\kmh{\frac{\mathrm{km}}{\mathrm{h}}}
\newglossaryentry{scene_tight_highway_entry_curve}{type=scenarios,
	name={\texttt{tight-\hspace{0pt}highway-\hspace{0pt}entry}},
    description={tight highway entry curve like section},
    user1={(real_20220608_142018, real_20220608_142523, real_20220608_143926)},
    user2={\ensuremath{60 \kmh}},
    user3={600},
	user4={\ensuremath{120\mathrm{s}}},
}
\newglossaryentry{scene_large_highway_entry_curve}{type=scenarios,
	name={\texttt{large-\hspace{0pt}highway-\hspace{0pt}entry}},
    description={large highway entry curve like section},
    user1={(bbg\_nord\_kurve)},
    user2={\ensuremath{100 \kmh}},
    user3={600},
	user4={\ensuremath{120\mathrm{s}}},
}
\newglossaryentry{scene_highway}{type=scenarios,
	name={\texttt{highway-\hspace{0pt}fast}},
    description={real highway scenario},
    user1={(real\_replay\_withlsadebug\_lane)},
    user2={\ensuremath{130 \kmh}},
    user3={1000},
	user4={\ensuremath{200\mathrm{s}}},
}
\newglossaryentry{scene_highway_slow_traffic}{type=scenarios,
	name={\texttt{highway-\hspace{0pt}slow}},
    description={real highway scenario with slow traffic},
    user1={(ORION\_65999)},
    user2={\ensuremath{75 \kmh}},
    user3={600},
	user4={\ensuremath{120\mathrm{s}}},
}
\usepackage{tikz}
\usetikzlibrary{shapes.symbols, arrows.meta, calc, angles, quotes} 

\newcommand{\nvar}[2]{%
    \newlength{#1}
    \setlength{#1}{#2}
}

\nvar{\evL}{0.875cm}
\nvar{\evW}{0.5cm}
\nvar{\laneW}{0.8cm}

\def\lane[#1,#2,#3]{%
   [bound line, double distance=\laneW, draw] #1 coordinate (end) (current subpath start) coordinate (start);
   \path [bound line, draw=white, line width=\laneW] (start) ++(#2:-1pt) -- #1 -- ++(#3:1pt);
   \path [center line, draw] #1
}

\tikzset{bound line/.style={solid, thin}}
\tikzset{center line/.style={dashed}}

\tikzset{
car/.style={car/.cd,#1, print},
    car/.cd,
    pos/.store in=\pos,
    pos=0.5,
    alpha/.store in=\alpha,
    alpha=100,
    latd/.store in=\latd,
    latd=0,
    angle/.store in=\angle,
    angle=0,
    args/.code={
      \tikzset{/tikz/car/argsstyle/.style={#1}}
    },
    prefix/.store in=\prefix,
    prefix=,
    argsstyle/.initial={},
    print/.style={
        /tikz/.cd,
        insert path={
            {
                coordinate[pos=\pos, sloped, shift={(0,\latd)}] (\prefix carpos) {}
                coordinate[pos=\pos, sloped, shift={(-.5\evL,\latd)}] (\prefix carmid) {}
                coordinate[pos=\pos, sloped, shift={(0,0)}] (\prefix carposref) {}
                coordinate[pos=\pos, sloped, shift={(-.5\evL,0)}] (\prefix carmidref) {}
                node[pos=\pos, anchor=west, draw=black!\alpha, sloped, shift={(-\evL,\latd)}, rotate around={\angle:(\prefix carpos)},
                minimum width=\evL, minimum height=\evW, 
                signal, signal to=east, signal pointer angle=90] {}
                node[pos=\pos, anchor=west, draw=black!\alpha, sloped, outer sep=2pt, shift={(-\evL-2pt,\latd)}, rotate around={\angle:(\prefix carpos)}, 
                minimum width=\evL, minimum height=\evW, car/argsstyle] (\prefix carnode) {}
            }
        }
    }
}

\title{\LARGE \bf
Trajectory Planning for Automated Driving using Target Funnels
}

\author{Benjamin Bogenberger$^{1}$, Johannes B\"urger$^{2}$, and Vladislav Nenchev$^{3}$
\thanks{${^1}$B.\;Bogenberger is with Technical University of Munich (TUM),  Munich, Germany (corresponding author) {\tt\small benjamin.bogenberger@tum.de}}%
\thanks{${^2}$J.\;B\"urger is with BMW Group, Munich, Germany {\tt\small johannes.buerger@bmw.de}}%
\thanks{${^3}$V.\;Nenchev is with the Institute of Embedded Systems, University of the Bundeswehr Munich, 85579 Neubiberg, Germany. {\tt\small vladislav.nenchev@unibw.de}}%
}

\begin{document}

\maketitle
\thispagestyle{empty}
\pagestyle{empty}

\begin{abstract}
Self-driving vehicles rely on sensory input to monitor their surroundings and continuously adapt to the most likely future road course. Predictive trajectory planning is based on snapshots of the (uncertain) road course as a key input. Under noisy perception data, estimates of the road course can vary significantly, leading to indecisive and erratic steering behavior. 
To overcome this issue, this paper introduces a predictive trajectory planning algorithm with a novel objective function: instead of targeting a single reference trajectory based on the most likely road course, tracking a series of target reference sets, called a target funnel, is considered. 
The proposed planning algorithm integrates probabilistic information about the road course, and thus implicitly considers regular updates to road perception.
Our solution is assessed in a case study using real driving data collected from a prototype vehicle.
The results demonstrate that the algorithm maintains tracking accuracy and substantially reduces undesirable steering commands in the presence of noisy road perception, achieving a 56\% reduction in input costs compared to a certainty equivalent formulation.



\end{abstract}
\glsresetall


\section{INTRODUCTION}

In recent years, car manufacturers have introduced increasingly sophisticated \acp{adas}, including hands-off driving, automatic lane changing and highly automated highway driving. However, many driving situations remain challenging because of the uncertainty in the vehicle's perceived environment. In this work we focus on perception uncertainties of the road course, which may be caused by the limited accuracy of map data or by limitations of the sensory input (e.g. camera). In self-driving applications of \ac{mpc}, snapshots of a continuously updated road course are passed on as a reference to an optimization-based trajectory planner which computes optimal control actions for the \ac{ev}. The reference trajectory is often obtained by reducing probabilistic road information to the most likely realization \cite{gutjahr_recheneffiziente_2019, liuPathPlanningAutonomous2017}. However, large updates of the most likely road course may cause indecisive and erratic steering behavior.

The problem setup bears similarities to classical robust and stochastic \ac{mpc} formulations \cite{mesbah_stochastic_2016, heirung_stochastic_2018, kouvaritakis_model_2016}. In these areas of research, uncertainty is typically introduced as either additive or multiplicative variations in the system model, with a primary focus on establishing control-theoretic properties such as robust stability \cite{Singh2017} and recursive feasibility \cite{Liniger2020}.
This is often achieved by minimax predictive control that optimizes the performance of the system with respect to the worst-case scenario \cite{Hans2014}. In \cite{lenz_stochastic_2015, Carvalho_2014} specific applications of \ac{smpc} to autonomous vehicles are considered in which the system model is uncertain. Additionally, uncertainty propagation is used to improve performance and feasibility in \ac{smpc}-based approaches for autonomous racing \cite{Zarrouki2024}. Optimization can also be carried out across the entire uncertainty space in an event-driven manner to enable robot exploration and control \cite{Nenchev2018}.
In contrast, in this paper we consider an uncertain reference trajectory derived from the measured road course. Previous work in process applications has addressed uncertain references within a known set \cite{ferramosca_mpc_2009,ferramosca_robust_2012}, where the system is controlled to a steady-state inside a given set by introducing an artificial steady-state and by penalizing the distance between the planned state and the artificial steady-state. In \cite{berger_funnel_2022}, an \ac{mpc} tracking is proposed for time-invariant nonlinear systems in the presence of a bounded disturbance that ensures that the state evolves in a prescribed performance funnel. Similarly, real-time motion plans that ensure safety despite environmental uncertainties were generated for a robot by pre-computing a library of funnels that define safe state boundaries during execution \cite{Majumdar2017}. However, the approach cannot be applied directly to automated driving, as it requires computing safe sets for the entire road network.

The main contributions of this paper include a probabilistic formulation of an \ac{mpc} problem with an uncertain reference trajectory, its approximation as a \ac{qp} problem using a target funnel that enables efficient computation in production systems with limited resources, and demonstrates its application to automated driving. Specifically, we focus on tracking a series of target reference sets defined at each point along the planning horizon, rather than relying on a single reference trajectory (i.e., a sequence of reference points).
Intuitively, this implies that within the target funnel there is no preferred state, which enhances the system's robustness to fluctuations in the road course. The provided case study demonstrates that variations in the input commands of a lateral motion planner are significantly reduced, thus improving comfort of the resulting automated driving system without compromising tracking accuracy and safety.
While our emphasis is on lateral trajectory planning, the approach is similarly applicable to longitudinal trajectory planning, where the uncertain reference trajectory is derived from a leading vehicle motion prediction.
Although the paper focuses on uncertainty in the tracking cost -- due to its most direct impact on planning performance in practice -- it is readily possible to extend the formulation by robust or stochastic constraints. In addition to the considered lane-keeping example, many real-world systems can utilize sensors to anticipate external signals in advance. Therefore, it can be expected that the proposed approach is applicable to a wide range of cyber-physical systems.

This paper is structured as follows: First, the problem statement is provided, and the stochastic \ac{ocp} is formulated (Sec.\,\ref{sec:problem_statement}). Then, the trajectory planning algorithm using target funnels in the objective function is presented (Sec.\,\ref{sec:solution}). Finally, in Sec.\,\ref{sec:eval} the approach is evaluated and compared with a baseline method for a lateral trajectory planner for automated driving using real data, followed by a discussion, and conclusions (Sec.\,\ref{sec:conclusions}).

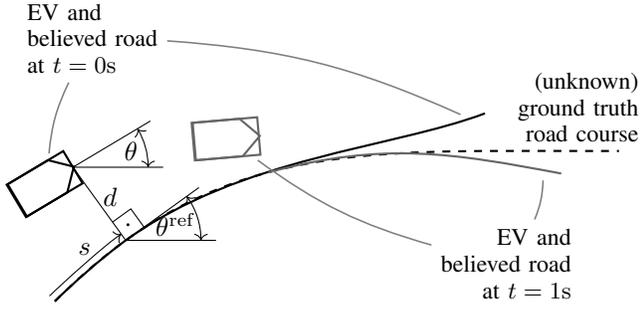
\begin{figure}
    \centering
    \def\myroadcourse{
(0,0) .. controls ++(45:3) and ++(0:-3) .. ++(7,2)
}

\tikzset{main/.style = {thick}}
\def\labelstrokeangle{10}

\pgfdeclarelayer{background}
\pgfdeclarelayer{foreground}
\pgfsetlayers{background,main,foreground}

\begin{tikzpicture}[semithick]
    \draw [main, dashed] \myroadcourse -- ++(0.5cm,0) node[label = {[inner sep=0pt, shift={(4mm,-1.5mm)}, anchor=south east, align=right, outer sep=4pt, font={\small}]above:{(unknown)\\ground truth\\road course}},pos=1] {};
    \path [main] \myroadcourse [car={pos=0.15,alpha=100,latd=1.2,angle=-5}];

    \begin{scope}[every path/.style={black}]
    \begin{pgfonlayer}{foreground}
        \draw (carposref) -- node[pos=0.6, right] {$d$} (carpos);
        
        \draw (carposref) -- ($(carposref) ! -1.2cm ! (carmidref)$) (carposref) -- ++(1.2cm, 0);
        \coordinate (angle0) at ($(carposref) + (1cm, 0)$);
        \coordinate (angle1) at ($(carposref) !-1cm! (carmidref)$);
        \pic [draw, <->,
          angle radius=10mm, angle eccentricity=0.7,
          "$\theta^{\mathrm{ref}}$"] {angle = angle0--carposref--angle1};
    
        \pic [draw,
          angle radius=3mm, angle eccentricity=.5,
          "."] {right angle = angle1--carposref--carpos};
    
        \draw[->] (0,0) ++(135:1mm) .. controls ++(45:0.2) and ++(30:-0.2) .. node[midway,above] {$s$} ($(carposref) ! 1mm ! (carpos)$);
    
        \draw (carpos) -- ($(carpos) !-1.2cm! (carnode.west)$) (carpos) -- ++(1.2cm, 0);
        \coordinate (angle0) at ($(carpos) + (1cm, 0)$);
        \coordinate (angle1) at ($(carpos) !-1cm! (carnode.west)$);
        \pic [draw, <->,
          angle radius=10mm, angle eccentricity=0.8,
          "$\theta$"] {angle = angle0--carpos--angle1};
    \end{pgfonlayer}
    \end{scope}

    \path \myroadcourse node[coordinate, pos=0.85, sloped, shift={(0, 0.5cm)}] (believedoffset) {};
    \path  \myroadcourse node[pos=0.15, sloped, shift={(1cm, 0)}] (control1) {}
                         node[pos=0.8, sloped, shift={(-1cm, 0)}] (control2) {};
    \draw [main] (0,0) .. controls ++(45:0.3) and ($(carposref)!-0.4!(control1)$) .. (carposref) (carposref) .. controls ($(carposref)!1.9!(control1)$) and ($(believedoffset)!0.9!(control2)$) .. (believedoffset) node[pos=0.95] (road) {};

    \node [align=left, font={\small}] (label1) at (0.5, 3.5) {EV and\\believed road\\at $t=0\mathrm{s}$};
    \draw [gray] (carnode) edge[bend left=\labelstrokeangle] (label1) (label1) edge[bend left=\labelstrokeangle] (road);
    
    \def\carposTwo{0.45}
    \path \myroadcourse [main, car={pos=\carposTwo,alpha=60,latd=0.5,angle=-10}];

    \path \myroadcourse node[coordinate, pos=0.97, sloped, shift={(0, -0.3cm)}] (believedoffset) {};
    \path  \myroadcourse node[pos=\carposTwo, sloped, shift={(1cm, 0)}] (control1) {}
                         node[pos=0.9, sloped, shift={(-1cm, 0)}] (control2) {};
    \draw [main, black!60] (carposref) .. controls ($(carposref)!1.6!(control1)$) and ($(believedoffset)!0.9!(control2)$) .. (believedoffset) node[pos=0.95] (road) {};

    \node [align=right, font={\small}] (label1) at (6, 0.5) {EV and\\believed road\\at $t=1\mathrm{s}$};
    \draw [gray] (carnode) edge[bend right=\labelstrokeangle] (label1) (label1) edge[bend right=\labelstrokeangle] (road);
\end{tikzpicture}
    \caption[Illustration of the EV and Its Observations of the Road Course]{The \ac{ev} in Frenet coordinates (shown are position $s$, absolute orientation $\theta$, and lateral displacement $d$ perpendicular to the reference with tangent angle $\theta^{\mathrm{ref}}$) and the \ac{ev}'s estimated road courses at times $t=0\mathrm{s}$ and $t=1\mathrm{s}$. Novel sensor information leads to updated road course estimates in between the two time steps, which can result in indecisive steering behavior. The road courses estimates are less accurate to the ground truth at greater distances.}\label{fig:illustrative_problem_and_frenet}
\end{figure}

\textbf{Notation.} The state and input at time $k \in \mathbb{Z}$ are denoted by $x_k \in \mathbb{R}^n$ and $u_k \in \mathbb{R}^m$, respectively. The stacked trajectory from time $k$ to $k+N$ is $\xTraj{k}{k+N} \coloneqq \begin{pmatrix} x_k^{\T} & x_{k+1}^{\T} & \cdots & x_{k+N}^{\T} \end{pmatrix}^{\T}$. Furthermore, we use $\indk{i}{\cdot}$ to express the $i$-step prediction of $\cdot$ given all the information available at time $k$, i.e., $\xk{i}$ and $\uk{i}$ are the predicted state and input, respectively. Analogously, we use $\xTraj[k]{k}{k+N}$ to denote a predicted trajectory.
The expected value is given by $\EB{\cdot}$. The probability of event $A$ is $\P(A)$. The element-wise $p$-th quantile of random the variable $x$ is $x_{(p)} = \min \left\{ 
	\big(z_1, \ldots, z_n \big)
	\mathop{:}
	\big(		\P(x_1 \leq z_1),
		\ldots,\P(x_n \leq z_n)
	\big) \geq p \right\}$.%
\section{Problem Statement}\label{sec:problem_statement}

We consider a vehicle model according to \cite{gutjahr_recheneffiziente_2019}, based on a linear approximation of the vehicle dynamics in Frenet coordinates, as illustrated in Fig.~\ref{fig:illustrative_problem_and_frenet}.
While vehicle dynamics are in general nonlinear \cite{mitschke_dynamik_2014}, it is possible to derive a linear approximation using contraction in Frenet coordinates \cite{gutjahr_recheneffiziente_2019}. This contraction involves two assumptions:
\begin{assumption}
    A small angle difference $\theta - \theta^{\mathrm{ref}}$ between \ac{ev} heading and road tangent, and a small ratio between lateral displacement and reference curve radius $\frac{d}{r^{\mathrm{ref}}} \ll 1$ are assumed.   
\end{assumption}
The resulting simplified dynamics are
    \begin{subequations}
    \begin{align}
        \dot d& = v (\theta - \theta^{\mathrm{ref}}) \,, \label{eq:lateral_simpler_dyn_a}\\
        \dot{\theta}&= v\kappa\,,
        \label{eq:lateral_simpler_dyn}
    \end{align}
    \end{subequations}
where $v$ and $\kappa$ denote the longitudinal velocity and curvature, respectively. The motion planning problem is divided into two sequential planning tasks: one in the longitudinal direction and the other in the lateral direction. During this process, one direction is fixed while the other is planned.
\begin{assumption}
As our focus is on lateral motion planning, we assume that the longitudinal state trajectory is predetermined and consider it as a known time-varying signal, as discussed in \cite{gutjahr_recheneffiziente_2019}.
\end{assumption}

The state for lateral motion planning contains the lateral displacement $d$, the absolute heading angle $\theta$, the curvature $\kappa$, and the derivative of the curvature with respect to time, i.e., $x=[d,\theta,\kappa,\dot{\kappa}]^\T\in \mathbb{R}^4$, and the control input $u_k \in \mathbb{R}$ is the second time derivative of the curvature (to allow for smooth steering). We discretize the lateral dynamics \eqref{eq:lateral_simpler_dyn} using the sampling time $\gls{ts} > 0$ \cite{gutjahr_recheneffiziente_2019} and thus obtain
\begin{equation}\label{eq:lat_dynamics}
    \begin{gathered}
    x_{k+1} = f_k(x_k, u_k, w_k) = \Mat A_k x_k + \Mat B_k u_k + \Mat D_k w_k \,, \\
    \Mat A_k = \begin{pmatrix}
        1 & v_k \gls{ts} & \frac{1}{2} v_k^2 \gls{ts}[^2] & \frac{1}{6} v_k^2 \gls{ts}[^3]  \\
        0 & 1            & v_k \gls{ts}                 & \frac{1}{2} v_k \gls{ts}[^2]    \\
        0 & 0            & 1                            & \gls{ts}                      \\
        0 & 0            & 0                            & 1                             \\
    \end{pmatrix} \,,\\
    \Mat B_k = \begin{pmatrix} 
        \frac{1}{24} v_k^2 \gls{ts}[^4] \\
        \frac{1}{6} v_k \gls{ts}[^3] \\
        \frac{1}{2} \gls{ts}[^2] \\
        \gls{ts}
    \end{pmatrix} \,,\quad
    \Mat D_k = \begin{pmatrix} 
        - v_k \gls{ts} \\
        0 \\
        0 \\
        0
    \end{pmatrix} \,.
    \end{gathered}
\end{equation}
For a safe operation of the \ac{ev} the state $x_k$ and input $u_k$ are constrained to the sets $\mathbb{X}$ and $\mathbb{U}$, respectively.
The disturbance $w_k = \theta^{\mathrm{ref}}_k$ (being equal to the absolute tangent angle of the road course provided by a perception module) follows from \eqref{eq:lateral_simpler_dyn_a}. It expresses how the lateral displacement depends on the angle error $\theta - \theta^{\mathrm{ref}}$. This disturbance acts as an uncertain reference for the system dynamics.

\subsection{Uncertain Reference and Disturbance}

The lateral reference trajectory is determined by the center line of the current lane and the longitudinal trajectory. In accordance with the lateral state definition the reference $\gls{ref}_k \in \mathbb{R}^n$ is defined as
\begin{equation}
    \gls{ref}_k = \begin{pmatrix}
            d^{\mathrm{ref}}_k &
            \theta^{\mathrm{ref}}_k &
            \kappa^{\mathrm{ref}}_k &
            \dot\kappa^{\mathrm{ref}}_k
    \end{pmatrix}^\T \;,
\end{equation}
where the desired lateral displacement $d^{\mathrm{ref}}$ is zero as we consider only cases where the \ac{ev} should drive in the lane center. 
However, the ground truth road course is not perfectly known. At each time step $k$ only an uncertain preview of the upcoming road is available. This preview is referred to as the road belief.
This belief can be represented by a conditional \ac{pdf} describing the knowledge about the road course at $i \geq 0$ steps in the future given all the information available at the current time instance $k$: The most likely believed road is represented by the expectation and the confidence in that belief is represented by the distribution. We denote this \ac{pdf} by $\obsk{i} \mathop{:} \mathbb{R}^n \mapsto [0,\infty)$ and use
\begin{equation}\label{eq:belief}
    \obsTraj[k]{k}{k+N} \mathop{:} \mathbb{R}^{(N+1) n} \mapsto [0,\infty)    
\end{equation}
for the belief \ac{pdf} of the road course from step $k$ to step $k+N$ (i.e., $\indTraj[]{k}{k+N}{\gls{ref}}$ the sequence of road courses over the planning horizon).
In the following we use $\indk{i}{\gls{ref}}$ and $\indTraj[k]{k}{k+N}{\gls{ref}}$ to denote the random variables distributed according to $\obsk{i}$ and $\obsTraj[k]{k}{k+N}$, respectively. 

\subsection{Stochastic Control Problem}

Combining the vehicle model \eqref{eq:lat_dynamics}, the state constraints, and the reference belief \eqref{eq:belief} yields the \ac{smpc} problem 
\begin{subequations}\label{eq:mpc}
    \begin{align}
        \argmin_{\uTraj[k]{k}{k+N-1}}&  \sum_{i=0}^{N} \norm*{\xk{i} - \indk{i}{\gls{ref}}}_{\Mat{Q}}^2 + \sum_{i=0}^{N-1} \norm{\uk{i}}_{\Mat{R}}^2 \label{eq:mpc_cost}\\
        \text{s.t.}\;\; & \xk{} =  x^p_k \\
        & \begin{multlined}[t]
            \xk{i+1} =  f_{k+i}( \xk{i}, \uk{i}, e_2^{\T} \indk{i}{\gls{ref}}) \\ i \in [0,N-1] 
        \end{multlined} \label{eq:mpc_dyn}\\
        & \xk{i} \in \mathbb{X} \quad i \in [0,N] \label{eq:mpc_state_cons}\\
        & \uk{i} \in \mathbb{U} \quad i \in [0,N-1] \label{eq:mpc_u_cons}\\
        & \indTraj[k]{k}{k+N}{\gls{ref}} \sim \obsTraj[k]{k}{k+N} \label{eq:mpc_belief}
    \end{align}
\end{subequations}
with standard basis vector $e_2$. $\Mat{Q}$ and $\Mat{R}$ are positive semidefinite and positive definite matrices, respectively, and $x^p_k$ denotes the initial state at a given time instant $k$. The \Ac{ocp} \eqref{eq:mpc} is stochastic due to the presence of the random variable in \eqref{eq:mpc_belief}. The \ac{cec} \cite{bertsekas_dynamic_1995} is an established approach for converting a stochastic \ac{ocp} to a deterministic \ac{ocp} and is considered as a baseline for comparison in this paper. While \eqref{eq:mpc} incorporates a reference trajectory, we interpret it as a planning problem with an emphasis on the tracking aspect. This perspective allows for the straightforward integration of additional objectives, such as obstacle avoidance as demonstrated in \cite{gutjahr_recheneffiziente_2019}.

\subsection{Certainty Equivalent Control Problem}\label{sec:cec}

The baseline \ac{cec} is based on an approximation of \Ac{ocp} \eqref{eq:mpc} by replacing all stochastic variables with their expected values. The substitution of $\indk{i}{\gls{ref}}$ with $\EB{\indk{i}{\gls{ref}}}$
leads to a deterministic \ac{ocp} with nominal state $\nk{i} = \EB{\xk{i}}$, which can be formulated as a \ac{qp} problem.

\section{UNCERTAINTY DEPENDENT TARGET FUNNELS}\label{sec:solution}

In this section, we introduce a novel solution to the \ac{smpc} \eqref{eq:mpc}. 
Unlike the \ac{cec} approach, which directly penalizes the distance to the expected value of the reference trajectory, we propose an alternative objective function that penalizes the distance relative to a target funnel (a time-varying set). This approach allows us to incorporate the uncertain information represented by the expected reference trajectory $\EB{\indTraj[k]{k}{k+N}{\gls{ref}}}$, while decreasing the sensitivity to fluctuations in that reference.

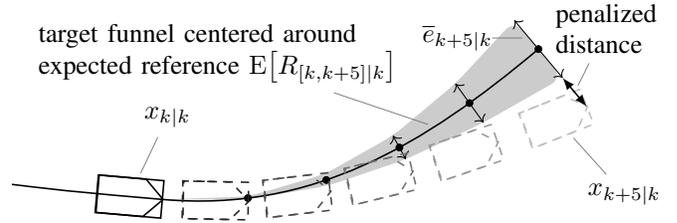
\begin{figure}
    \centering
    \def\myroadcourse{
(-2,0.2) .. controls ++(-7:0.8) and ++(-5:-1) .. (0,0)
}

\def\myroadcourseA{
(0,0) .. controls ++(-5:2) and ++(40:-2) .. ++(5,2)
}

\def\poslist{0.2, 0.4, ..., 1}      
\def\countlist{1,2,...,5}           
\def\countlistrev{5,4,...,1}        
\def\countlistminusone{1,2,...,4}   
\def\countlistexp{3,4,5}            
\def\elabelat{5}                    

\begin{tikzpicture}[semithick]
    \path \myroadcourseA \foreach \x [count=\xi] in \poslist
        {node[coordinate, pos=\x,sloped,shift={($\x*\x*(0,0.5)$)}] (a\xi) {}};
    \path \myroadcourseA \foreach \x [count=\xi] in \poslist
        {node[coordinate, pos=\x,sloped,shift={($\x*\x*(0,-0.5)$)}] (b\xi) {}};
    \path[fill=gray!40, rounded corners=5pt] (0,0) \foreach \xi in \countlist {-- (a\xi)} \foreach \xi in \countlistrev {-- (b\xi)} -- cycle;
    
    \draw [black!100] \myroadcourse \myroadcourseA [car={pos=0,
    args={pin={$\xki[k]{}$}}
    }] node[pos=0.7,inner sep=0pt,outer sep=2pt,pin={[align=left]130:{target funnel centered around\\expected reference $\EB{\indTraj[k]{k}{k+5}{\gls{ref}}}$}}] {};
    \path \myroadcourseA \foreach \x [count=\xi] in \poslist 
        {node[circle, fill, inner sep=1pt, pos=\x, ] (c\xi) {}};

    \draw \foreach \xi in \countlistexp {(a\xi) edge[<->, thin] (b\xi)};
    \path (a\elabelat) -- node[pos=0.7, inner sep=0pt, outer sep = 2pt, pin={[pin distance=4mm, inner sep=0pt, outer sep = 0pt]180:$\indk{5}{\overline{e}}$}] {} (c\elabelat);

    \path \myroadcourseA \foreach \x [count=\xi] in \poslist 
        {node[circle, fill, inner sep=1pt, pos=\x, ] (c\xi) {}};

    \path \myroadcourseA \foreach \x [count=\xi] in \poslist
        {node[coordinate, pos=\x,sloped,shift={($\x*\x*(0,-1)$)}] (a\xi) {}};
    \path [dashed] (0,0) \foreach \xi in \countlistminusone {
    -- (a\xi)
    let \n1 = {100-15*\xi} in
    [car={pos=1, alpha=\n1}]
    }
     -- (a\elabelat)
    let \n1 = {100-15*\elabelat} in
    [car={pos=1, alpha=\n1,args={pin={[pin edge=solid]below:$\xki[k]{5}$}}}];
    \draw (a\elabelat) edge[arrows = {latex-latex}] (b\elabelat);
    \path (a\elabelat) -- node[pos=0.5, inner sep=0pt, outer sep = 2pt, pin={[align=left, pin distance=3mm,anchor=230]85:penalized\\distance}] {} (b\elabelat);
\end{tikzpicture}
    \caption[Interpretation of the \acl*{method_uts} Approach]{The uncertainty dependent Target Funnel approach: The \ac{ev} at $\xki[k]{}$, the expectation of its belief reference and an exemplary motion plan (dashed) are shown. The black dots depict the discretization points/times. The gray area illustrates the target funnel whose size $\overline e$ corresponds to the belief's uncertainty (only lateral displacement dimension is shown).}
    \label{fig:best_case_shifting_illustration}
\end{figure}

The target funnel is constructed such that at prediction steps $i$ with high confidence beliefs the target funnel cross sections are small while at prediction steps with low confidence beliefs the target funnel cross sections are large. In Fig.~\ref{fig:best_case_shifting_illustration} this target funnel is depicted by the gray area and the penalized distance at prediction instance $5$ between a possible plan and the funnel is shown by the black arrow. All states within the target funnel are equally desirable for the motion of the automated driving vehicle. 
First, we show how the objective function can be modified to penalize the distance to a target funnel.

\subsection{Penalizing the Distance to a Set}\label{sec:objective_set_distance}

The target funnel $\indTraj[k]{k}{k+N}{\mathcal{R}}$ can be interpreted as a sequence of target reference sets $\indk{i}{\mathcal{R}}$ each centered on a specific point on the nominal reference trajectory $\nk{i}$. Additionally, the target funnel itself is time varying (i.e., it will vary from one time instant $k$ to the next depending on updated information about the road course). We propose to modify the objective \eqref{eq:mpc_cost} such that it minimizes the squared euclidean distance between state trajectory and a hyperrectangular set $\indTraj[k]{k}{k+N}{\mathcal{R}}$. This distance is given by
 \begin{equation}\label{eq:hyperbox_distance}
    \min_{r \in \indTraj[k]{k}{k+N}{\mathcal{R}}} \norm{\nTraj[k]{k}{k+N}-r}^2 \,,
\end{equation}
rather than by the distance between the nominal state trajectory $\nTraj[k]{k}{k+N} \in \mathbb{R}^{(N+1)n}$ and the reference trajectory $\indTraj[k]{k}{k+N}{r} \in \mathbb{R}^{(N+1)n}$, given by $\norm{\nTraj[k]{k}{k+N} - \indTraj[k]{k}{k+N}{r}}^2$.
\begin{definition}
Let us define the target funnel as a hyperrectangular set
\begin{equation}\label{eq:def_target_funnel}
\indTraj[k]{k}{k+N}{\mathcal{R}} = \{ r \in \mathbb{R}^{(N+1)n} \mathop{:} \abs{r - \indTraj[k]{k}{k+N}{C}} \leq \frac{\indTraj[k]{k}{k+N}{\overline{e}}}{2} \}    
\end{equation}
with center $\indTraj[k]{k}{k+N}{C} \in \mathbb{R}^{(N+1)n}$ and edge lengths $\indTraj[k]{k}{k+N}{\overline{e}} \in \mathbb{R}^{(N+1)n}$, whose derivation will be discussed in the next section.
\end{definition}
Inserting \eqref{eq:def_target_funnel} into \eqref{eq:hyperbox_distance} yields \begin{equation}\label{eq:hyperbox_distance_final}
    \begin{gathered}
        \min_{\indTraj[k]{k}{k+N}{r}} \sum_{i=0}^{N} \norm{\nk{i} - \indk{i}{r}}^2 \\
        \text{s.t.}\, \abs{\indTraj[k]{k}{k+N}{r} - \indTraj[k]{k}{k+N}{C}} \leq \frac{\indTraj[k]{k}{k+N}{\overline{e}}}{2} \,,
    \end{gathered}
 \end{equation}
where we use the fact that $\indTraj[k]{k}{k+N}{\mathcal{R}}$ is a hyperrectangle. In order to apply the target funnel to the original \ac{smpc} \eqref{eq:mpc}, we determine the target funnel parameters in the following.

\subsection{Target Funnel Definition}\label{sec:target_set_def}

To apply the concept of a tracking cost function using a target funnel as defined in \eqref{eq:hyperbox_distance_final}, the target funnel will be defined based on the road belief. Specifically, we select the edge lengths of the funnel such that it contains the $\gls{parameter-p}\%$ most likely realizations of the stochastic tracking error $\indk{i}{e} = \xk{i} - \indk{i}{\gls{ref}}$ (in an element-wise sense). The parameter $\gls{parameter-p} \in [0, 1)$ therefore determines how far the funnel extends outwards from the expected reference.

For normally distributed environment beliefs this corresponds to the case that the target funnel is centered on the expected reference trajectory
\begin{subequations}
\begin{equation}
    \indTraj[k]{k}{k+N}{C} = \EB{\indTraj[k]{k}{k+N}{\gls{ref}}} \,,
\end{equation}
and the widths $\indk{i}{\overline{e}}$ of the target funnel are defined by
\begin{equation}\label{eq:target_set_size}
    \indk{i}{\overline{e}} = e_{k+i|k, \left(\frac{1}{2} + \frac{\gls{parameter-p}}{2}\right)} - e_{k+i|k, \left(\frac{1}{2} - \frac{\gls{parameter-p}}{2}\right)} \quad 0 \leq \gls{parameter-p} < 1 \,,
\end{equation}
\end{subequations}
where \glslink{quantile}{$\cdot_{(p)}$} denotes the element-wise $p$-th quantile. For $\gls{parameter-p}=0$ the target funnel width is zero and the objective function equals the \ac{cec} objective function, for $\gls{parameter-p} \rightarrow 1$ the funnel size approaches infinity which implies that the tracking cost tends to zero.
Note that, while the tracking error $\indk{i}{e}$ itself depends on the input, the width $\indk{i}{\overline{e}}$ depends only the variance of the reference trajectory.

\subsection{Optimal Control Problem based on Target Funnels}

\begin{figure}
    \centering
    \includegraphics{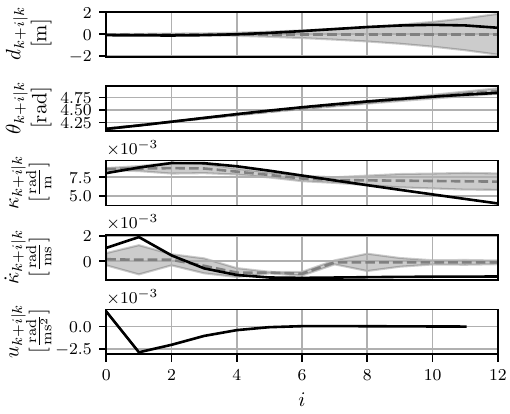}
    \caption{Exemplary planning step with the target funnel method showing the planned trajectory (black) and the uncertainty dependent target funnel (gray) (in scenario \gls*{scene_tight_highway_entry_curve}).}\label{fig:example_planning_step}
\end{figure}

We now consider the \ac{smpc} \eqref{eq:mpc} with two modifications: crucially, we first apply the definition of the target funnel \eqref{eq:hyperbox_distance_final} to the objective function, and, secondly, we consider a nominal evolution of the model dynamics (similarly to 
the \ac{cec}). The proposed trajectory planning method based on a target funnel can thus be stated as:
\begin{subequations}\label{eq:tfunnel}
    \begin{align}
        \argmin_{\uTraj[k]{k}{k+N-1}}& \min_{\indTraj[k]{k}{k+N}{r}} \sum_{i=0}^{N} \norm*{\nk{i} - \indk{i}{r}}_{\Mat{Q}}^2 + \sum_{i=0}^{N-1} \norm{\uk{i}}_{\Mat{R}}^2     \label{eq:target_set_cost} \\
        \text{s.t.}\;\; & \abs*{\indTraj[k]{k}{k+N}{r} - \EB{\indTraj[k]{k}{k+N}{\gls{ref}}}} \leq \frac{\indTraj[k]{k}{k+N}{\overline{e}}}{2} \\
        & \nk{} =  x^p_k \label{eq:tfunnel_initial}\\
        & 
        \begin{multlined}[t]
            \nk{i+1} =  f_{k+i}( \nk{i}, \uk{i}, e_2^{\T} \EB{\indk{i}{\gls{ref}}}) \\ i \in [0,N-1] 
        \end{multlined} \label{eq:tfunnel_dyn}\\
        & \nk{i} \in \mathbb{X} \quad i \in [0,N] \label{eq:tfunnel_state_cons}\\
        & \uk{i} \in \mathbb{U} \quad i \in [0,N-1] \label{eq:tfunnel_u_cons}\\
        & \indTraj[k]{k}{k+N}{\gls{ref}} \sim \obsTraj[k]{k}{k+N} \label{eq:tfunnel_belief}
    \end{align}
\end{subequations}
where \eqref{eq:tfunnel_initial}-\eqref{eq:tfunnel_belief} follow from \eqref{eq:mpc}. This method introduces $(N+1)n$ additional box constrained decision variables compared to the \ac{cec}. \Ac{mpc} \eqref{eq:tfunnel} is deterministic because it relies solely on the expected value of the random variable in \eqref{eq:tfunnel_belief}, allowing efficient solution as a \ac{qp} with off-the-shelf solvers. An exemplary target funnel and the resulting planned trajectory are depicted in Fig.~\ref{fig:example_planning_step}.
\section{EVALUATION}\label{sec:eval}

In this section, we evaluate the proposed planning algorithm that uses target funnels. It is important to note that the primary input -- the perceived road course -- is generated from real driving data collected during prototype vehicle drives on a test track. The performance metrics are assessed through \ac{cl} lane-keeping simulations.

\subsection{Scenario Simulation}
The simulations are conducted with real driving and perception data obtained from four highway scenarios (approximate velocity noted in parentheses): 1) \gls{scene_tight_highway_entry_curve} (\glsentryuserii{scene_tight_highway_entry_curve}), 2) \gls{scene_large_highway_entry_curve} (\glsentryuserii{scene_large_highway_entry_curve}), 3) \gls{scene_highway_slow_traffic} (\glsentryuserii{scene_highway_slow_traffic}) and 4) \gls{scene_highway} (\glsentryuserii{scene_highway}).

\begin{figure}
    \centering
    \subfigure[\gls{scene_tight_highway_entry_curve}]{
        \includegraphics{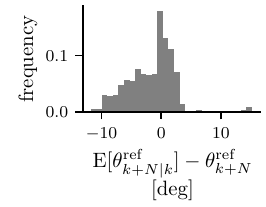}
    }%
    \subfigure[\gls{scene_highway}]{
        \includegraphics[trim={0.7cm, 0, 0, 0},clip]{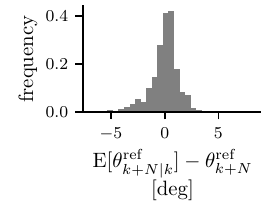}
    }
    \caption{Histograms over all time steps $k$ of the estimation error of the road tangent angle at the end of the prediction horizon.}\label{fig:error_hist}
\end{figure}

\begin{figure}
    \vspace{0.1cm} 
    \centering
    \def\ltrim{0.8cm}
    \def\toptrim{0cm}
    \def\lowertrim{0.5cm}
    \subfigure[\gls{scene_tight_highway_entry_curve}]{
        \includegraphics[trim={0, \lowertrim, 0, \toptrim},clip]{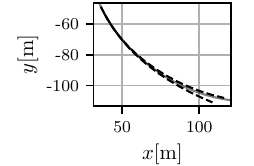}
    }%
    \subfigure[\gls{scene_large_highway_entry_curve}]{
        \includegraphics[trim={\ltrim, \lowertrim, 0, \toptrim},clip]{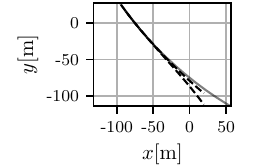}
    }
    
    \def\lowertrim{0}
    \subfigure[\gls{scene_highway_slow_traffic}]{
        \includegraphics[trim={0, \lowertrim, 0, \toptrim},clip]{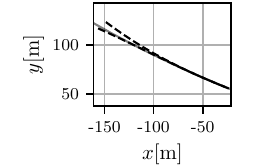}
    }%
    \subfigure[\gls{scene_highway}]{
        \includegraphics[trim={\ltrim, \lowertrim, 0, \toptrim},clip]{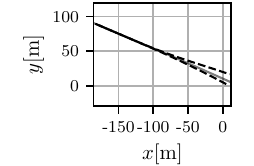}
    }
    \caption{Perceived road course (dashed) at two consecutive time steps and ground truth road (gray).}\label{fig:scenarios}
\end{figure}

To illustrate the uncertainty in the perception data, we show the error distribution between the perceived road tangent angle $\E[\indTraj[k]{k}{k+N}{\theta^{\mathrm{ref}}}]$ and the ground truth tangent angle $\indTraj[k]{k}{k+N}{\theta^{\mathrm{ref}}}$ in Fig.~\ref{fig:error_hist}. Snapshots of road estimates at two consecutive time steps are depicted in Fig.~\ref{fig:scenarios} illustrating how the perceived road course varies from one time step to the next.

Algorithm~\ref{alg:simulation} outlines how we reprocess each recorded scenario. Each simulation step consists of: Acquiring the recorded road belief and resampling this belief at positions given by the recorded longitudinal trajectory (lines \ref{algline:perception}, \ref{algline:resamplebelief}). Then, our proposed \ac{mpc} \eqref{eq:tfunnel} is used to generate the lateral plan (line \ref{algline:lateralplan}) and the simulation is advanced (line \ref{algline:stepsystem}). Finally, we compute the performance metrics (line \ref{algline:computemetrics}), which are introduced next.

\begin{algorithm}
  \caption{Reprocessing procedure with computation of performance metrics\label{alg:simulation}}
  \begin{algorithmic}[1]
    \Statex \textbf{input:} $s$:~Scenario, $x^{\mathrm{lon}}_{[\cdots]}$:~Recorded longitudinal trajectory, $\xki[0]{}$:~Initial lateral state
    \Let{$K$}{\Call{ScenarioIterations}{$s$}}
      \For{$k \gets 0 \textrm{ to } K-1$}
            \Let{$\gls{obs}_{\mathrm{view range}| k}$}{\Call{Perception}{$\indk{}{x^{\mathrm{lon}}} , s$}} \label{algline:perception}
            \Let{$\indTraj[k]{k}{k+N}{\gls{obs}}$}{\Call{Resample}{$\gls{obs}_{\mathrm{view range}| k} , \indTraj{k}{k+N}{x^{\mathrm{lon}}}$}} \label{algline:resamplebelief}
            
            \vspace{0.5\baselineskip}
            \Let {$\xTraj[k]{k}{k+N}, \uTraj[k]{k}{k+N-1}$}{\Call{Plan}{$\xk{}, \indTraj[k]{k}{k+N}{\gls{obs}}$}} \label{algline:lateralplan}
            \Let {$\xki[k+1]{}$}{$\xk{1}$} \label{algline:stepsystem}

            \vspace{0.5\baselineskip}
            \Let {$\gls{ref}_{k+1}$}{\Call{GroundtruthRoad}{$x^{\mathrm{lon}}_{k+1} , s$}} 
            \State \Call{ComputeMetrics}{$\xki[k+1]{}, \uk{}, \gls{ref}_{k+1}$} \label{algline:computemetrics}
      \EndFor
  \end{algorithmic}
\end{algorithm}

Note that we use the planning horizon length $N = 12$ with sampling time $\gls{ts}=0.5\mathrm{s}$, the cost weights $\Mat Q = \eye$ and $\Mat R = 100$, and the target funnel parameter $\rho = 60\%$. The admissible state and input sets are $\mathbb{X} = \left\{ x_k\in \mathbb{R}^n \mathop{:} \abs{\kappa} \leq 0.02 \frac{\mathrm{rad}}{\mathrm{m}} \right\}$ and $\mathbb{U} = \left\{ u \in \mathbb{R} \mathop{:} \abs{u} \leq 0.425 \frac{\mathrm{rad}}{\mathrm{m}\mathrm{s}^2} \right\}$, respectively. To  investigate the impact of the funnel more explicitly, we disregarded lane boundaries in our problem formulation. In practice, the funnel may be constrained by lane boundaries in the immediate vicinity of the vehicle, where perception quality is typically adequate, and further constraints as in \cite{gutjahr_recheneffiziente_2019} may be readily integrated to guarantee safety. Constraints may also be slacked to guarantee feasibility.

\subsection{Performance Metrics}

In practice, applying the \ac{cec} to our problem setting with updated road beliefs at each time step (see Sec.~\ref{sec:cec}) has resulted in erratic steering movements in the considered driving scenarios. The goal of this work is to reduce these critical variations in the steering actuation compared to the \ac{cec} without compromising the tracking accuracy.

We note that the two terms of the stage cost in the \ac{ol} objective function \eqref{eq:mpc_cost} capture these goals. By evaluating these two terms with the \ac{cl} trajectory with respect to the ground truth road course and normalizing them over the entire scenario length $K$ we obtain our two metrics:
\begin{itemize}
    \item The \textbf{\ac{cl} deviation cost} measuring the tracking performance (assuming the ground truth road $\gls{ref}_k$ is known)
    \begin{subequations}
    \begin{equation}\label{eq:cl_deviation_cost}
        J^{\vec x} \coloneqq \frac{1}{K+1} \sum_{k=0}^{K} \norm{\xk{} - \gls{ref}_k}_{\Mat Q}^2 \;,
    \end{equation}
    \item the \textbf{\ac{cl} input cost} measuring the steering movements
    \begin{equation}\label{eq:cl_input_cost}
        J^{\vec u} \coloneqq \frac{1}{K} \sum_{k=0}^{K-1} \norm{\uk{}}_{\Mat R}^2 \;.
    \end{equation}
    \end{subequations}
\end{itemize}

\subsection{Results}

\begin{figure}
    \vspace{0.1cm} 
    \centering
    \includegraphics[trim={0, 0.6cm, 0, 0.6cm},clip]{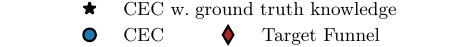}
    \centering
    \includegraphics[trim={0, 0.3cm, 0, 0.05cm},clip]{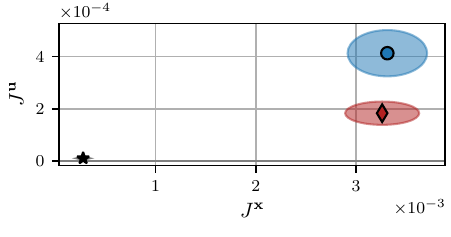}
    \caption{Mean \ac{cl} deviation cost and \ac{cl} input cost over all scenarios. The semi-major/-minor axis of the ellipses are half the standard deviation of the respective metric.}\label{fig:best_N12_100_2d}
\end{figure}

\begin{figure}
    \vspace{0.1cm} 
    \centering
    \includegraphics[trim={0, 0.1cm, 0, 0.1cm}, clip]{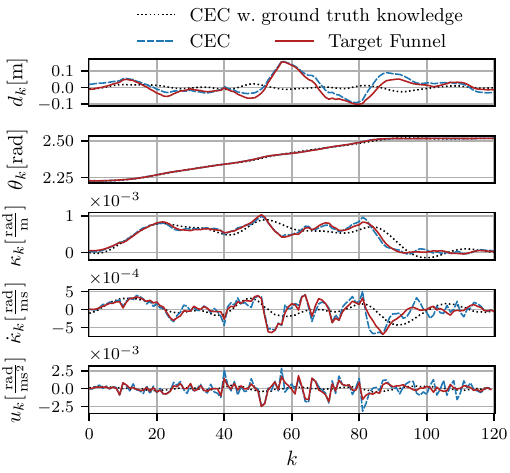}
    \caption{Exemplary section of the \ac{cl} trajectories resulting from the different planning methods in the scenario \gls{scene_highway_slow_traffic}.}\label{fig:cl_trajectory}
\end{figure}

Figure~\ref{fig:best_N12_100_2d} shows the mean and standard deviation of the performance metrics over all scenarios. Exemplary sections of the resulting \ac{cl} trajectory are shown in Fig.~\ref{fig:cl_trajectory}.
Firstly, in both figures we can observe a significant degradation of the CL performance under uncertainty compared to the idealized \ac{cec} with ground truth knowledge (black) illustrating the nature of the problem: Oscillating input commands lead to an undesirable driving experience.
However, in Fig.~\ref{fig:best_N12_100_2d} we see that the proposed target funnel method reduces the mean \ac{cl} input cost by $56\%$ compared to the \ac{cec}, while improving marginally the tracking performance. Further, the improvements can also be observed in \ac{cl} trajectory (Fig.~\ref{fig:cl_trajectory}): the target funnel method prevents large input command spikes (bottom chart, especially at $k \in [100, 120]$), while the resulting state trajectories (top four charts) of the \ac{cec} and target funnel approach are very similar.

%
\subsection{Discussion} \label{sec:discussion}

As demonstrated in the preceding evaluation, our method adapts well to the current perception uncertainty. Under uncertain road knowledge the planning algorithm based on a target funnel results in significantly less unnecessary and indecisive steering movements compared to the \ac{cec} (mean input costs are reduced by up to $56\%$ with marginally improved tracking accuracy). We demonstrated that the reduction in steering movement is visually noticeable in exemplary trajectories, suggesting that its effect would be noticeable for a passenger of the autonomously driving \ac{ev}. We thus expect that our method improves comfort in the automated vehicle. At the same time computational efficiency is ensured, since only the solution of a \ac{qp} with moderately increased complexity compared to the \ac{cec} is needed. This makes the method deployable for \ac{adas} solutions in production with limited computational performance.

Tuning certainty equivalent \ac{mpc} is known to be challenging because of the need for robustness against disturbances, an appropriate selection of prediction horizons, and the balancing of various weighting factors. By reducing the effects of disturbances, the proposed method is expected to simplify tuning, particularly in combination with automatic tuning approaches such as \cite{Wu2024}.

Finally, we envision that our method can be readily applied in other cyber-physical systems, where sensors are used to obtain uncertain limited previews of external signals.%
\section{CONCLUSION}\label{sec:conclusions}

This paper presented a computationally efficient \ac{mpc} trajectory planner for automated driving that accounts for road course uncertainty using a target funnel. The method improves comfort while maintaining tracking performance and safety despite uncertain and continuously updated road information as shown in a case study using real driving data.

Future work will assess the effect on comfort in closed-loop driving in a real vehicle. We also aim to explore uncertainty in state constraints, address theoretical guarantees such as stability, and investigate longitudinal or combined longitudinal/lateral trajectory planning with uncertain references,
such as a leading vehicle with perception noise and prediction uncertainty.
%




\bibliographystyle{ieeetr} 
\bibliography{bib}

\end{document}